\documentclass[runningheads]{llncs}
\usepackage{graphicx}
\usepackage{amsmath,amssymb} 
\usepackage{color}
\usepackage[width=122mm,left=12mm,paperwidth=146mm,height=193mm,top=12mm,paperheight=217mm]{geometry}
\usepackage{bm}
\usepackage{subfig}

\newcommand{\yoshi}[1]{\textcolor{black}{#1}}
\newcommand{\yoshia}[1]{\textcolor[rgb]{0.0,0.0,0.0}{#1}}
\newcommand{\rei}[1]{\textcolor{black}{#1}}
\newcommand{\reia}[1]{\textcolor[rgb]{0.0,0,0.0}{#1}}
\newcommand{\fukuda}[1]{\textcolor{black}{#1}}
\newcommand{\fukudaa}[1]{\textcolor{black}{#1}} 
\newcommand{\allen}[1]{\textcolor[rgb]{0.0,0.0,0.0}{#1}}

\begin{document}
\pagestyle{headings}
\mainmatter
\def\ECCV18SubNumber{1269}  

\title{Cross-connected Networks for Multi-task Learning of Detection and Segmentation} 




\author{Seiichiro Fukuda$_1$
\hspace{2mm} Ryota Yoshihashi$_1$ 
\hspace{2mm} Rei Kawakami$_1$
\\ \hspace{6mm}Shaodi You $_{2,3}$
\hspace{10mm}Makoto Iida$_1$
\hspace{5mm}Takeshi Naemura$_1$}
\institute{1 The University of Tokyo 
\hspace{2mm} 2 Data61-CSIRO
\hspace{2mm} 3 Australian National University}

\maketitle

\begin{abstract}
Multi-task learning improves generalization performance by sharing knowledge among related tasks. Existing models are for task combinations annotated on the same dataset, while there are cases where multiple datasets are available for each task. How to utilize knowledge of successful single-task CNNs that are trained on each dataset has been explored less than multi-task learning with a single dataset. We propose a cross-connected CNN, a new architecture that connects single-task CNNs through convolutional layers, which transfer useful information for the counterpart. We evaluated our proposed architecture on a combination of detection and segmentation using two datasets. Experiments on pedestrians show our CNN achieved a higher detection performance compared to baseline CNNs, while maintaining high quality for segmentation. It is the first known attempt to tackle multi-task learning with different training datasets between detection and segmentation. Experiments with wild birds demonstrate how our CNN learns general representations from limited datasets.
\keywords{Multi-task Learning, Pedestrian Detection, Bird Detection, Semantic Segmentation, Convolutional Neural Networks}
\end{abstract}

\section{Introduction}\label{sec:introduction}
Multi-task learning 
aims to improve the generality of performance by mutually utilizing information of other \rei{related} tasks \cite{caruana1997multitask}. 
By modeling multiple tasks in a single network, useful knowledge among tasks can be shared during training, and diversified training data tend to cancel out bias and noise.

\begin{figure}[t]
\renewcommand{\arraystretch}{0.8}
	\begin{center}
    {\tabcolsep=2.0mm
    \begin{tabular}{ccc}
    \includegraphics[width=0.27\hsize]{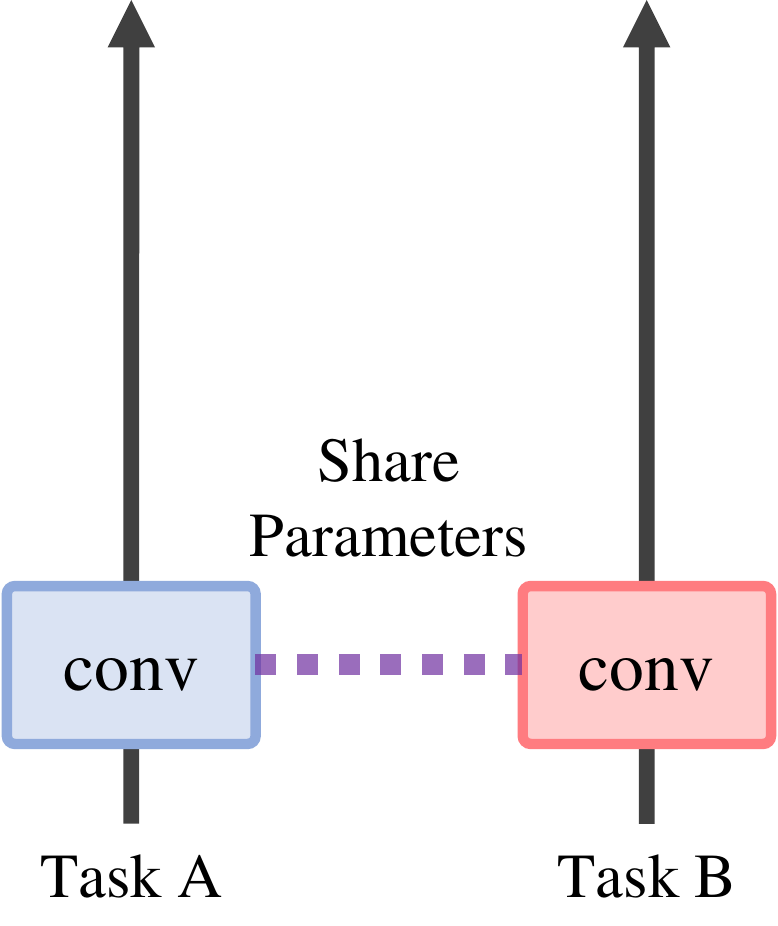} \label{subfig:compare_paramshare} &
    \includegraphics[width=0.27\hsize]{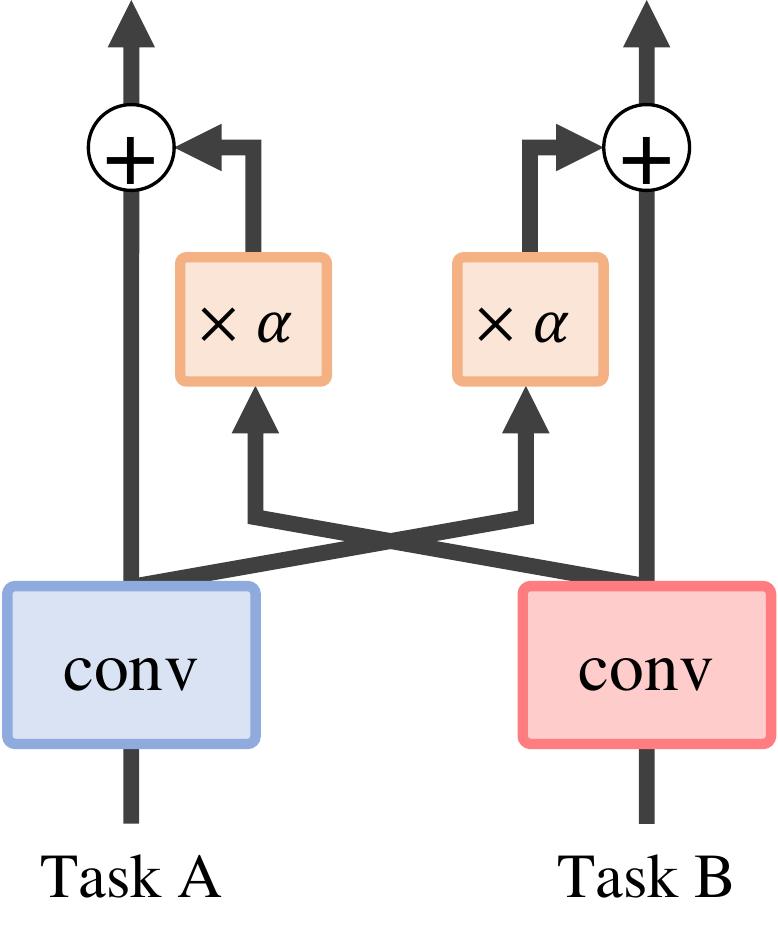} \label{subfig:compare_cross-stitch} &
    \includegraphics[width=0.27\hsize]{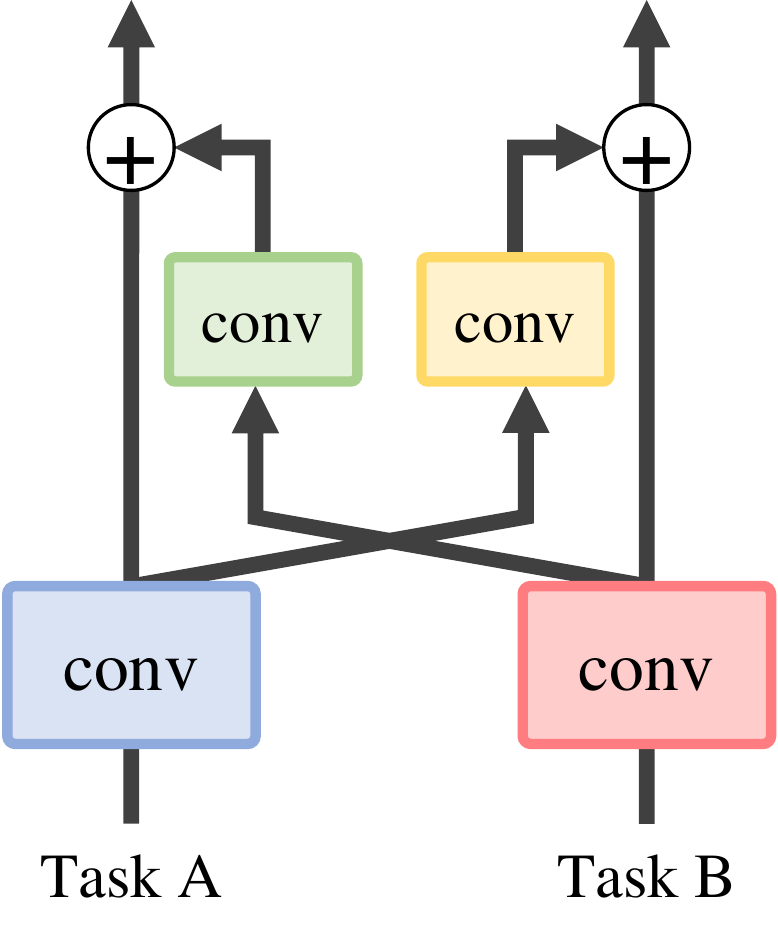} \label{subfig:compare_ours} \\
    (a) Parameter sharing &
    (b) Cross-stitching \cite{misra2016cross} &
    (c) Cross connection (ours)\\
    \end{tabular}
    }
    \end{center}
\renewcommand{\arraystretch}{1.0}
    \vspace{-3mm}
    \caption{Comparison of basic units that constitute multi-task CNNs. (a) Parameter sharing: Two layers learn the same representations suitable for both tasks. 
    (b) Cross-stitching: Two scaling layers model channel-wise weighted sum of activation maps from conv layers. (c) Cross connection (ours): Two cross-connecting layers model linear combination of activation maps utilizing all channels.}
    \label{fig:network_comparison}
    \vspace{-7mm}
\end{figure}

The most common way to achieve 
multi-task learning is to share parameters in feature representation
layers of single-task convolutional neural networks (CNNs)
and branch several top layers for task-wise prediction \cite{bischke2017multi,eigen2015predicting,tian2015pedestrian,kao2016visual,zhang2014facial,kendall2017multi,gkioxari2014r} \rei{as illustrated in} Fig.\,\ref{fig:network_comparison}\,(a).
The shared layers, either by hard or soft sharing~\cite{ruder2017overview}, learn common representations for all tasks, and they achieve better generalization because more data are harder to overfit. 
However, this sharing architecture is not very flexible, since sharing choices are discrete, and the number of shared layers must be fixed among all tasks.
Particularly in upper layers, the parameter sharing can be too restrictive because feature representations must be specialized on each task~\cite{zeiler2014visualizing,yosinski2014transferable}.

To alleviate this, cross-stitch networks~\cite{misra2016cross} were proposed as a more general CNN architecture for multi-task learning. 
As shown in Fig.\,\ref{fig:network_comparison}\,(b), 
it models shared representations by using element-wise linear combinations of activation maps from each task stream,
while retaining individual network parameters.
It is, however, limited in that it considers only the combination of channels with corresponding indices.
Cross-stitching has been applied to multi-tasks that have annotations on the same dataset, but as we tend to have different learning datasets for different tasks, we may like to use successful single-task CNNs trained on each dataset, and combine them later for better performance.

In this paper, we propose a cross-connected CNN, a new architecture for multi-task CNNs.
As shown in Fig.\,\ref{fig:network_comparison}\,(c),  
we cross-connect intermediate layers of 
\reia{single-task} 
CNNs via convolutional layers. Our architecture \allen{enables} 
\reia{task-wise} 
streams to communicate with each other by exchanging their activation maps, 
while its novelty is in that the activation map passes through convolutional layers.
The convolution layers learn the importance of each activation map for the other task, and 
\allen{determines} which information to be sent to which destination.
The proposed architecture is a generalization of cross-stitching, and it can further model mutual effects across channels.
The cross-connection can be made across any layers in principle.

\fukuda{To verify the effectiveness of cross-connected CNN, we compare its performance with that of several baseline CNNs.
\reia{For two} tasks to combine, we employ object detection and semantic segmentation, 
\reia{a combination that can be trained on independent dataset.}
Considering properties of their annotations, they may benefit from multi-task learning.
Bounding boxes for object detection have a potential to help semantic segmentation to localize a specific kind of object.
Pixel-level labels over an entire image for semantic segmentation have a potential to provide knowledge for holistic understanding of a target scene, which may encourage robust detection in front of complex backgrounds.
\yoshia{While existing studies in multi-task learning only use datasets that provide both detection and segmentation annotations~\cite{teichmann2016multinet,dvornik2017blitznet}, we tackle multi-task learning from different datasets of two tasks to exploit more diverse information sources. }
\fukudaa{In the experiments, we examined whether the cross-connected CNN can learn useful representations,}
\reia{using two types of objects that are common to the two tasks: pedestrians and wild birds.}}
\fukudaa{Experiments on pedestrians show the proposed CNN architecture produces better detection performance by leveraging knowledge of segmentation even when the training datasets are different between tasks.
In the experiments on wild birds, our CNN achieved a higher generalization performance compared to baselines.}

\vspace{2mm}
\noindent{\bf Contributions \hspace{2mm}}
This paper has the following contributions.
(1) We propose a cross-connected CNN, a new architecture for multi-task learning.
Convolutional layers that cross-connect two single-task CNNs can model cross-channel and cross-layer feature interaction between tasks, and the proposed model is a generalization of existing ones. 
(2) To our knowledge, this is the first attempt to tackle multi-task learning of object detection and semantic segmentation using different datasets between tasks. 

\vspace{-2mm}
\section{Related Work}
\vspace{-2mm}
Multi-task learning is a method to divert useful knowledge of one task to other tasks. 
\yoshia{Although multi-task learning with simple parameter sharing~\cite{caruana1997multitask} is successful 
in various tasks~\cite{bischke2017multi,eigen2015predicting,tian2015pedestrian,zhang2014facial,kao2016visual,gkioxari2014r},}
\yoshia{the combinations of tasks are based on one of the following assumptions to ease parameter sharing: 
First, 
one task is an auxiliary task to the other, such as pose estimation and action recognition\,\cite{gkioxari2014r}, and facial landmark detection and attribute prediction\,\cite{zhang2014facial}.
Second, 
one task is short of training data, and thus helped by the annotations of the other task, as in depth estimation and surface normal prediction\,\cite{eigen2015predicting}. 
Finally, 
both tasks have the same training sets with multiple labels, which is valid for all of the above examples.}
\fukudaa{In these cases, 
hard parameter sharing in hidden 
layers is flexible enough to successfully learn shared representations. } 
\yoshia{However, more flexible networks are preferable to enable combination of broader tasks that do not meet the assumptions.}

\fukudaa{Toward more flexible multi-task networks, soft parameter sharing \cite{duong2015low,yang2017trace} and cross-stitch networks \cite{misra2016cross} have been proposed, 
and both of them combine pre-trained single-task CNNs to utilize their knowledge.
Soft parameter sharing regularizes single-task parameters and encourages them to be similar. Its network structure is still within the paradigm of parameter sharing.
The cross-stitch networks have more flexible structure. They model linear combinations of activation maps between two task streams, 
and the parameters in single-task streams as well as in connections can vary to fit the data without restrictions.
However, the cross-stitch networks are limited in that the combinations of activation maps are restricted to corresponding channels.
We show that our cross-connected CNN effectively utilizes two single-task CNNs pre-trained on different datasets between tasks: object detection and semantic segmentation.}

\fukudaa{\yoshia{Focusing on multi-task learning of object detection and semantic segmentation,} there are a few studies utilizing deep CNNs \cite{teichmann2016multinet,kokkinos2016ubernet,dvornik2017blitznet}.
MultiNet \cite{teichmann2016multinet} incorporates three single-task CNNs for classification, detection and segmentation by parameter sharing.
UberNet \cite{kokkinos2016ubernet} and BlitzNet \cite{dvornik2017blitznet} aggregates activation maps from middle layers of a single CNN via task-wise skip connections, as similarly done in\,\cite{doersch2017multi}, but 
\reia{since all the task-wise streams \fukudaa{use} the activation maps from the 
same single CNN, \fukudaa{they are} within} the classical paradigm of parameter sharing.
We differ from \cite{teichmann2016multinet,kokkinos2016ubernet,dvornik2017blitznet} in two respects.
First, we construct a multi-task CNN by integrating two single-task CNNs pre-trained on each task.
Although consuming a lot of memory due to increased network parameters, our CNN can easily reuse existing network architectures.
Second, we use different training datasets between tasks in our evaluation, while in \cite{teichmann2016multinet,kokkinos2016ubernet,dvornik2017blitznet}, they are trained on the same dataset.
}


\fukuda{Instance segmentation \cite{dai2015convolutional,pinheiro2015learning,dai2016instance,hu2018learning} is also a candidate task to \fukudaa{combine} with object detection, which can distinguish individual object areas \fukudaa{of} the same class.
However, fewer annotations for instance segmentation are available than \allen{those} for semantic segmentation, since it \allen{requires} even more annotation effort. \yoshi{Learning from partial annotation \cite{hu2018learning} can mitigate this labor, but at the \allen{cost} of segmentation accuracy.}
In this paper, we \allen{chose} semantic segmentation for multi-task learning with object detection.}

\yoshia{Apart from multi-task learning, late-fusion-based output refinement~\cite{cheng2017segflow,kalogeiton2017joint} is a promising approach to simultaneously improve multiple outputs 
from deep networks for multiple tasks. This type of methods is useful for correlated predictions, such as segmentation and optical flow~\cite{cheng2017segflow}, and object and action detection~\cite{kalogeiton2017joint}. Those methods are different from this paper in the motivations, as their aim is to improve performance by integrating two correlated outputs, while ours is to improve generalization of feature-level representations. }


\vspace{-2mm}
\section{Cross-connected CNN}
\begin{figure}[t]
	\begin{center}
		\includegraphics[width=\hsize]{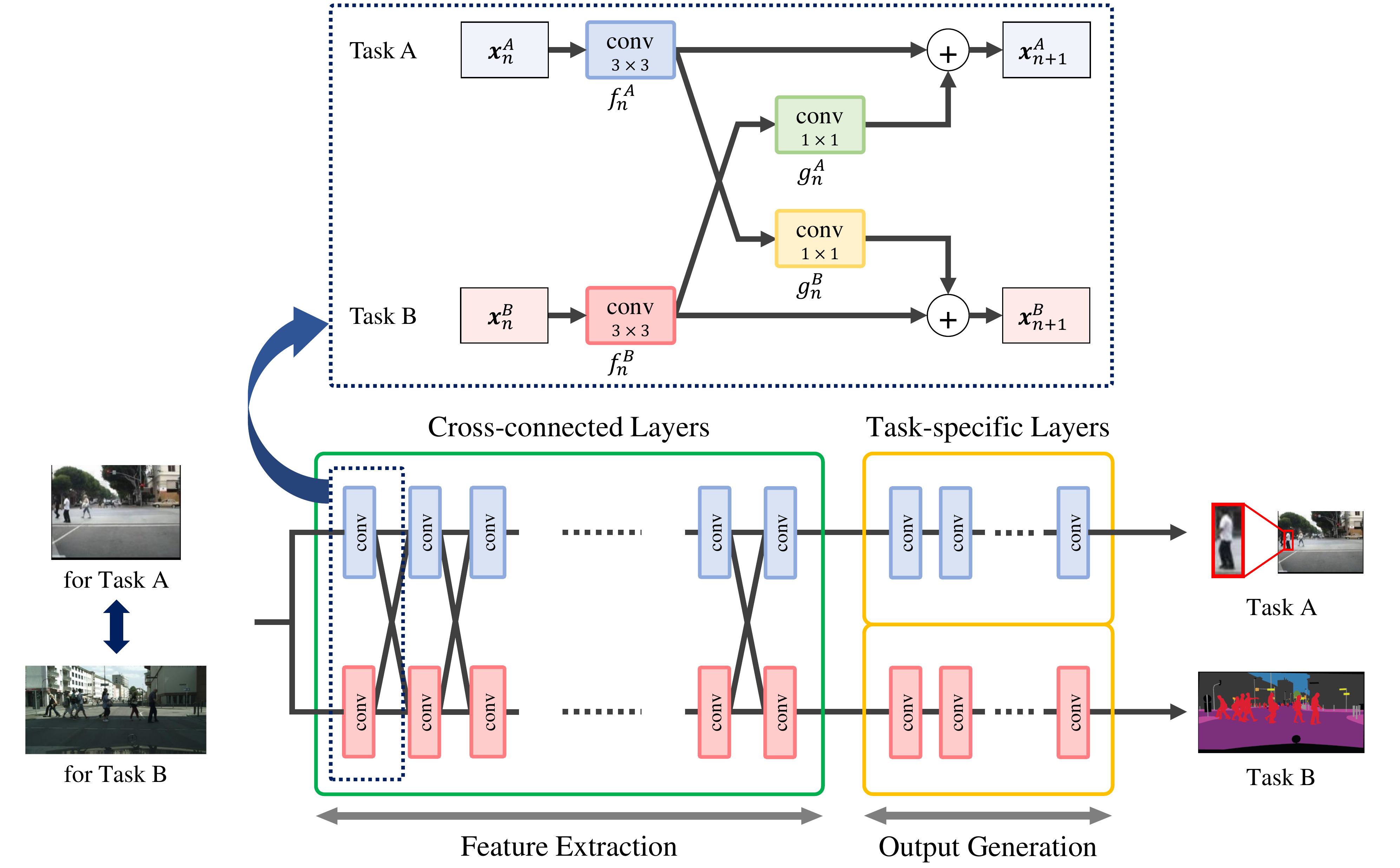}
	\end{center}
    \vspace{-5mm}
    \caption{Overview of cross-connected CNN: We integrate two single-task CNNs for Task A (blue) and Task B (red) by \allen{using} cross connections.
Cross-connected layers are common to both tasks.
${1\times1}$ convolutional layers on cross connections (green and yellow) \fukudaa{model} linear transformation and \fukudaa{mutually} transmit useful information \fukudaa{between tasks}.
Task-specific layers \allen{generate outputs for each task separately on the basis of} feature maps extracted in cross-connected layers. \fukudaa{All conv blocks are composed of a convolutional layer and an activation function ReLU.}}
	\label{fig:network_overview_and_unit}
    \vspace{-5mm}
\end{figure}

\vspace{-2mm}
We propose cross-connected convolutional networks for multi-task learning.
\reia{In the following, we explain our method by taking two-task learning as an example. 
We denote the two tasks as Task A and Task B.}
As shown in Fig. \ref{fig:network_overview_and_unit}, \reia{we cross-connect feature extraction layers of two single-task CNNs via convolutional layers.} 
Our network consists of cross-connected layers shared by both tasks, and task-specific layers separated for each task.\\

\vspace{-2mm}
\subsection{Components of Cross-connected CNN}
\vspace{-1mm}
\noindent{\bf Cross-connected Layers \hspace{2mm}}
Cross-connected layers are designed to \fukudaa{effectively} share knowledge between \fukuda{the combined tasks}.
\allen{The cross-connected layers are represented} as a stack of the basic unit as illustrated in 
\reia{a dotted rectangle} in Fig.~\ref{fig:network_overview_and_unit}.
It shows how the ${n}$-th unit receives input maps and passes output maps to the ${n+1}$-th unit. The unit consists of original convolutional layers derived from single-task CNNs (drawn in blue and red) and additional convolutional layers connecting two CNNs (drawn in green and yellow). 
\fukudaa{All convolutional layers are followed by an activation function ReLU (omitted for simplicity in the figure).}
\reia{The connecting convolutional layers have \allen{as many ${1\times1}$ kernels as} the number of channels in the output maps of the other stream. The kernel size is chosen to be one-by-one, since the connection is for knowledge transfer rather than feature extraction. The connections represent linear transformation, and learn the importance of each activation map for the other task.}
\reia{We denote input maps for $n$-th unit as ${\bm{x}_{n}^{A}}$ and ${\bm{x}_{n}^{B}}$, and the transformations learned by \allen{the} original convolutional layers \fukudaa{and ReLU} as ${f_{n}^{A}}$ and ${f_{n}^{B}}$.}
\reia{Assuming cross-connection layers \fukudaa{and ReLU} learn transformations ${g_{n}^{A}}$ and ${g_{n}^{B}}$, then ${\bm{x}_{n+1}^{A}}$ and ${\bm{x}_{n+1}^{B}}$ are computed as}
\begin{eqnarray}
	\begin{cases}
		\bm{x}_{n+1}^{A} = f_{n}^{A}(\bm{x}_{n}^{A}) + g_{n}^{A}(f_{n}^{B}(\bm{x}_{n}^{B})) & \\
		\bm{x}_{n+1}^{B} = f_{n}^{B}(\bm{x}_{n}^{B}) + g_{n}^{B}(f_{n}^{A}(\bm{x}_{n}^{A})). &
	\end{cases}
	\label{eqn:multi-task_output}
\end{eqnarray}
Activation maps are added \yoshi{in element-wise manner}.
The second terms ${g_{n}^{A}(f_{n}^{B}(\bm{x}_{n}^{B}))}$ and ${g_{n}^{B}(f_{n}^{A}(\bm{x}_{n}^{A}))}$ have information considered useful for one task based on the knowledge obtained in the other task.
\fukudaa{In the first unit, both ${\bm{x}_{1}^{A}}$ and ${\bm{x}_{1}^{B}}$ are equal to an input RGB image.}

\vspace{2mm}
\noindent{\bf Task-specific Layers \hspace{2mm}}
\fukudaa{Task-specific layers are prepared for each task and trained to be specialized on task-wise output generation.
For example, object detection branches its network into two paths, which are responsible for bounding box regression and its classification, respectively.
Semantic segmentation generates a map with pixel-wise labels.
Since these layers are expected to perform different functions between tasks, their architectures are also required to be designed differently.
Therefore, generally, they cannot be cross-connected due to the difference of the shape of their activation maps.
Task-specific layers take feature maps from cross-connected layers and separately process them without communication between tasks.}

\subsection{Training Procedure}\label{subsec:training_procedure}
The training procedure of cross-connected CNNs consists of \allen{two steps: single-task} and multi-task learning.\\

\noindent{\bf Single-task Learning \hspace{2mm}}
We \yoshi{first pre-train} CNNs for \fukuda{each task} independently with individual datasets, \yoshia{without cross connections}.
\yoshi{By pre-training single task networks}, we can utilize task-specific knowledge of one task for the other task during multi-task learning more easily.
\allen{To cross-connect them}, we have to \allen{select} CNNs with a common structure.
\yoshi{Each CNN is trained by minimizing} task-specific loss functions: \fukuda{${L_A}$ for Task A and ${L_B}$ for Task B}.\\ 

\noindent{\bf Multi-task Learning \hspace{2mm}}
\yoshi{\allen{Having pre-trained single-task CNNs}, we start to train the cross-connected network. Layers \allen{in the network} are initialized by weights from pre-trained single-task networks,
except \allen{for} cross-connecting convolutional layers\allen{, which are} inserted after pre-training. The cross-connecting convolutional layers are initialized by random weights, and \fukudaa{they learn} to transform and transfer activation maps of one task to the others after \allen{being} updated by multi-task training.}

As in \cite{kokkinos2016ubernet}, we use the sum of task-specific losses as a multi-task loss.
We denote it as ${L_{all}}$ \allen{that} satisfies the following expression:
\begin{equation}
	L_{all} = L_{A} + \lambda L_{B}.
	\label{eqn:multi_loss}
\end{equation}
\yoshi{\allen{All} layers in the network are updated in an end-to-end manner \allen{in accordance with} the gradient of both loss functions.}
\yoshi{To enable multi-task learning in different datasets between tasks, we switch training datasets at a constant interval.}
We compute a loss of only one task and set \allen{the} loss of \allen{the other} task that has no annotations to zero.
That is, \fukuda{${L_{all}=L_{A}}$ for Task A, and ${L_{all}=\lambda L_{B}}$ for Task B}.

\section{Experimental Evaluation}
To examine the effect of a cross-connected CNN, we compare its performance with single-task CNNs as well as 
those of the existing models on multi-task learning of object detection and semantic segmentation. 
In those tasks, we show that detection is enhanced by rich contextual information from segmentation, and semantic segmentation is more aware of important targets by detection.
Specifically, we present experiments on two domains: pedestrians and wild birds.

\subsection{Network Implementation}
We first prepare two single-task CNNs, each of which is specialized for object detection and semantic segmentation.
We use region proposal network (RPN) \cite{zhang2016faster} based on VGG16 \cite{simonyan2014very} for detection, which has been shown to be effective in pedestrian recognition.
\fukuda{As in \cite{zhang2016faster}, we use a smooth L1 loss for bounding box regression and a cross-entropy loss for its classification}.

For semantic segmentation, we construct
a VGG16-based pyramid scene parsing network (PSPNet) \cite{zhao2017pyramid} 
by combining the convolutional layers from VGG16 and a pyramid pooling module \cite{zhao2017pyramid}.
Although the original PSPNet is based on ResNet, we use VGG16-based ones to clarify the effect of multi-task learning.
We use a cross-entropy loss summed over all pixels.
Those two single-task CNNs are also used as baselines for comparison.

Having the two VGG16-based single-task CNNs, we connect them to construct a cross-connected CNN.
We cross-connect the first 10 convolutional layers (conv1\_1--conv4\_3) and assign the rest as task-specific layers.
We set ${\lambda}$ in Eq. \ref{eqn:multi_loss} as 1.0, following related studies \cite{teichmann2016multinet,dvornik2017blitznet}.
We fine-tune the cross-connected network by the training procedure in Sec. 3.2. 
Since our architecture has more parameters than the single-task CNNs, performance improvement may be natural.
To clarify the improvement by multiple datasets from that by more number of parameters, 
we also fine-tune cross-connected CNNs only with a single dataset as baselines.
They are denoted as `single-task cross-connected', and it is fine-tuned only by detection dataset when evaluating detection, 
and by segmentation dataset for segmentation evaluation.

For the multi-task baselines to compare with ours, we implement two types of multi-task CNNs, 
one is hard parameter sharing and the other is a cross-stitch network~\cite{misra2016cross}. Both of them are applied to VGG16-based RPN and PSPNet, and trained in the procedure in Sec. 3.2.
For the parameter sharing, we test four types of CNNs, each of them
shares layers up to the 1st (2 layers), the 2nd (4 layers), the 3rd (7 layers), and the 4th pooling layer (10 layers), respectively.
We refer to each of them as `Share1,' `Share2,' `Share3,' and `Share4,' on the basis of the index of the \fukuda{top} pooling layer. 
A cross-stitch network, denoted as `Cross-stitch,' is implemented by replacing convolutional layers in the cross-connected CNN
with scale layers, which learn channel-wise multiplicative scaling factors.

\subsection{Evaluation Metrics}
We use log-average Miss Rate on False Positives Per Image (FPPI) within a defined range to evaluate detection performance.
For semantic segmentation, we use Intersection over Union (IoU), which is used to evaluate the segmentation accuracy of target areas.

In addition to the common ones, we define Detection Rate, a metric focused on the recognition performance of a specific class, 
as the number of detected instances divided by the total number of instance regions:
\begin{equation}
\text{Detection Rate} = \frac{\#\text{(detected instances)}}{\#\text{(total instances)}}.
\end{equation}
This measure aims to evaluate awareness of instances, which should be enhanced by the jointly-learned detection. 
It is introduced since pixel-wise measures such as IoU do not penalize totally ignored instances. 
A higher Detection Rate indicates more instance regions are correctly labeled. 
Each instance is considered to be detected if the detected area meets the following condition: 
$\frac{area(gt \cap pred)}{area(gt)} \ge S_{th},$
where ${gt}$ and ${pred}$ denote the ground truth and prediction labels of an instance, and ${S_{th}}$ is a threshold.
We take the average of Detection Rate when changing ${S_{th}}$ from 0.1 to 0.9 in increments of 0.1.

\subsection{Pedestrian Detection and Segmentation}\label{subsec:pedestrian_recognition}
We first evaluate our proposed architecture by using pedestrian detection and segmentation.
In the field of road-scene understanding, while a recent dataset~\cite{Cordts2016Cityscapes} provides full pixel-wise labels, many datasets including the largest one~\cite{dollar2012PAMI} only have person bounding boxes. 
Having CNNs trained on each large dataset, we verify whether the cross-connected CNN successfully improves performance in both tasks. 
As the detection dataset only has `person' labels, we focus on `person' as a specific foreground object class in segmentation.\\

\noindent{\bf Datasets \hspace{2mm}}
We use Caltech Pedestrian \cite{dollar2012PAMI} for detection, a video dataset of urban road scenes taken at ${640\times480}$ pixels.
Pedestrians in each frame are annotated with bounding boxes.
According to the official assignment from~\cite{dollar2012PAMI}, we used 42,782 images (set00--set05) for training and 4,024 images (set06--set10) for testing.
For the test, we use the `reasonable' evaluation subset which targets only pedestrians sized at 50 pixels or taller and at least 65\% visible.
Log-average Miss Rate is calculated on FPPI in ${[10^{-2}, 10^0]}$ after filtering the proposal boxes by non-maximum suppression (NMS) with a threshold of 0.7.

For semantic segmentation, we use Cityscapes \cite{Cordts2016Cityscapes}, a recently released dataset for semantic understanding of urban road scenes.
It consists of 5,000 images with ${2,048\times1,024}$ pixels annotated with pixel-level labels.
The images are divided into 2,975 for training, 500 for validation, and 1,525 for testing.
We use the validation set for \fukudaa{testing} because annotations for the original test set are available only in the official evaluation server.
\fukudaa{19-class labels are assigned to each pixel and we focus on `person' and `rider' \allen{as} target classes.}
In this experiment, we integrate the two labels for label consistency with Caltech Pedestrian.
\\

\noindent{\bf Training Details \hspace{2mm}}
We train multi-task CNNs in accordance with the procedure described in Sec.~\ref{subsec:training_procedure}.
For detection, RPN\,\cite{zhang2016faster} pre-trained on ImageNet\,\cite{russakovsky2015imagenet} is fine-tuned on Caltech Pedestrian.
As in \cite{zhang2016faster}, we first resize images in Caltech to ${960\times720}$ pixels,
and then crop 4 patches with ${512\times384}$ pixels from them, due to GPU memory constraints.
For semantic segmentation, VGG16-based PSPNet pre-trained on ImageNet and PASCAL VOC is fine-tuned on Cityscapes.
We randomly crop patches with ${512\times384}$ pixels from images in Cityscapes.
Convolutional layers in cross connections are initialized with Gaussian distribution with the standard deviation of 0.1.
Since we use different datasets between tasks, we switch datasets during training at an interval of 100 iterations.\\

\begin{figure}[t]
	
\end{figure}

\noindent{\bf Results of Pedestrian Detection \hspace{2mm}}
\begin{figure}[t]
	\vspace{-4mm}
	\begin{center}
		\includegraphics[width=0.8\hsize]{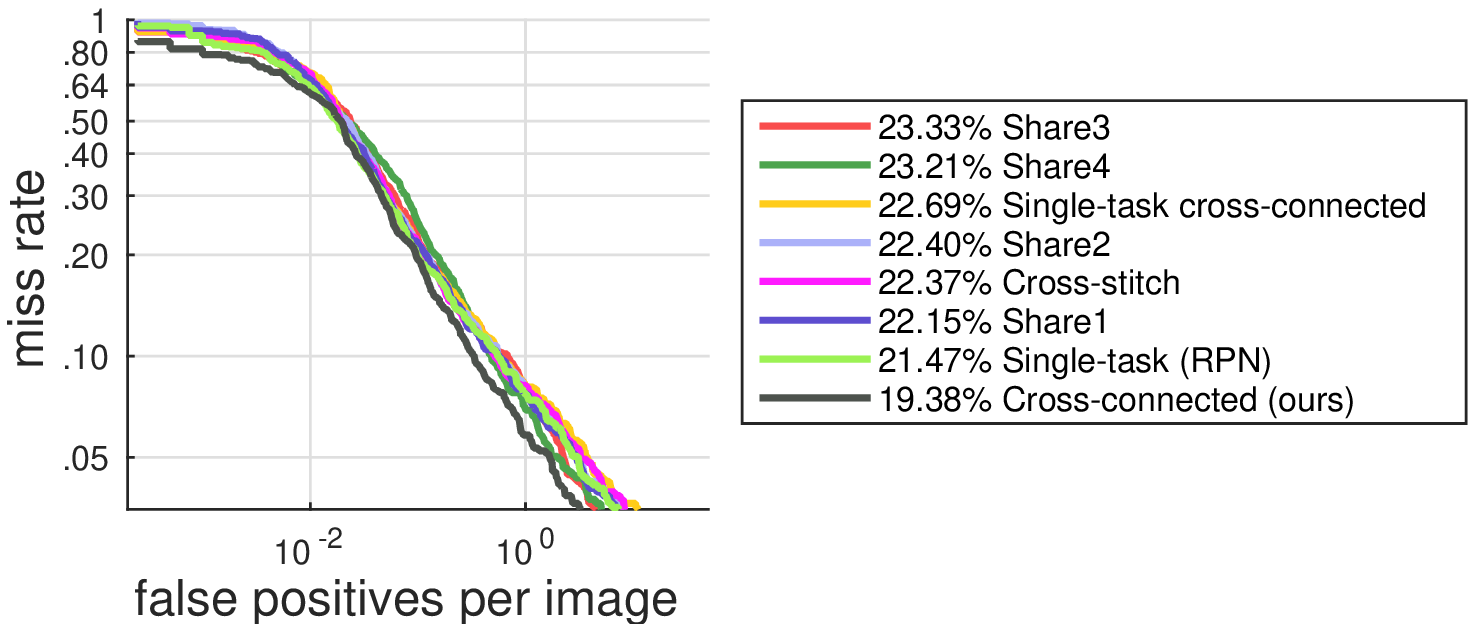}
	\end{center}
    \vspace{-5mm}
    \caption{Evaluation results of pedestrian detection on Caltech Pedestrian. The legend indicates log-average Miss Rate.}
	\label{fig:caltech_results}
    \begin{center}
		\includegraphics[width=\hsize]{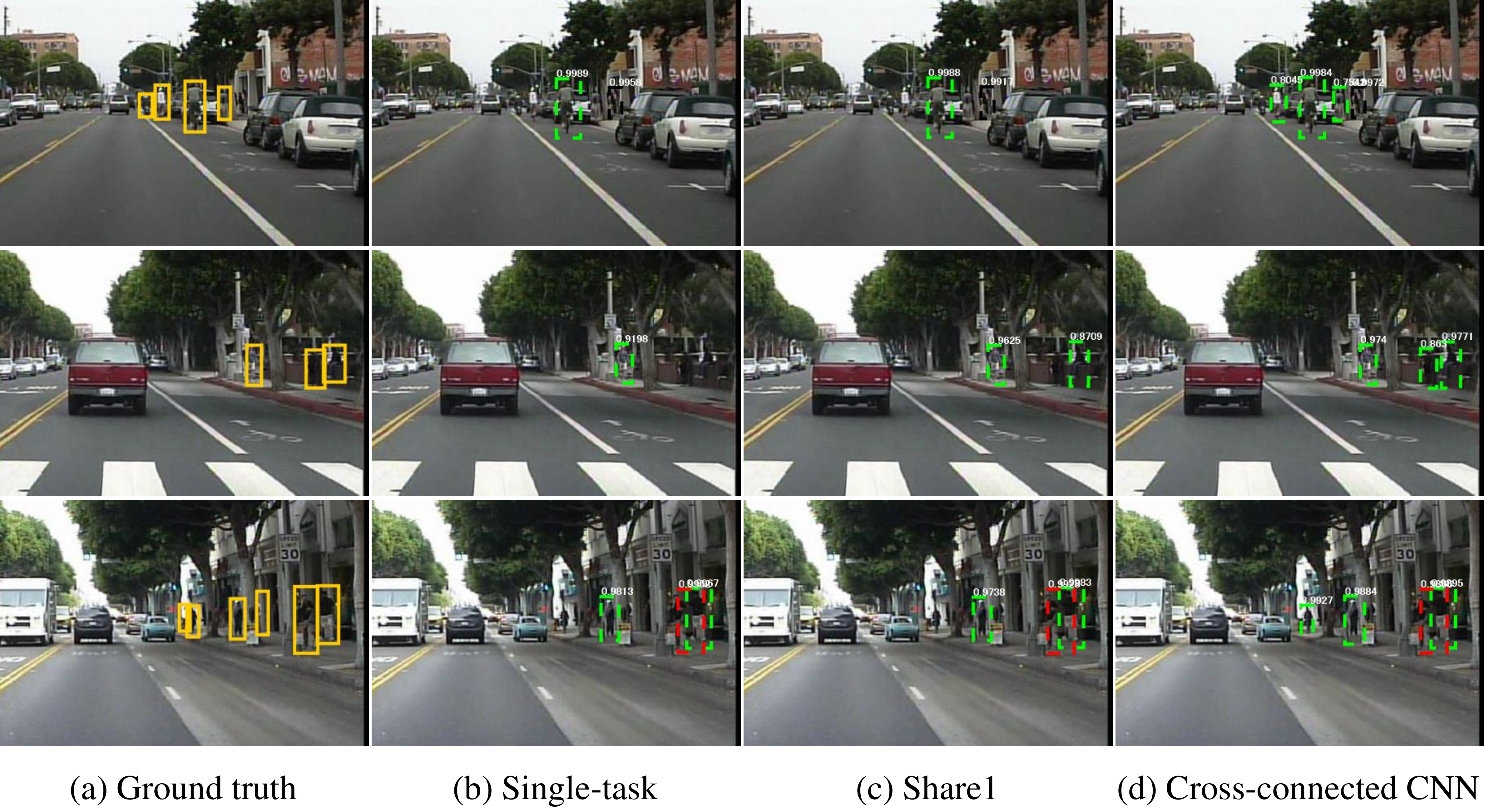}
	\end{center}
    \vspace{-5mm}
    \caption{Example detections on Caltech Pedestrian by baselines and cross-connected CNN. The ground truth, true positives, and false positives are displayed in yellow, green, and red rectangles, respectively.}
	\label{fig:caltech_examples}
    \vspace{-7mm}
\end{figure}
Fig. \ref{fig:caltech_results} shows the evaluation results on Caltech Pedestrian.
Multi-task baselines (`Share1--4' \fukuda{and `Cross-stitch'}) have log-average Miss Rate\allen{s} of 22.15--23.33\%, 
which are worse by 0.68--1.86\% compared to that of `Single-task'.
This suggests that hard parameter sharing suffers from less flexibility, 
and the cross-stitch network fails in utilizing the single-task knowledge because interaction of activation maps is limited between corresponding channels.
On the other hand, the cross-connected CNN achieves log-average Miss Rate of 19.38\%, which is 2.09\% better than `Single-task'.
In addition, it is 3.31\% better than `\fukudaa{Single-task cross-connected}'.
The cross-connected CNN successfully leverages the knowledge of semantic segmentation even when it is trained by different datasets for each task.
Fig.\,\ref{fig:caltech_examples} shows example detections \allen{using} the single-task CNN, multi-task baseline (`Share1'), and cross-connected CNN.
The cross-connected CNN can detect more pedestrians \allen{that are} relatively hard to recognize in front of complex backgrounds.\\

\noindent{\bf Results of Segmentation} \hspace{2mm}
\begin{table}[t] 
	\begin{center}
    	\caption{\fukuda{Results of semantic segmentation on Cityscapes}}
        {\tabcolsep = 4mm
    	\begin{tabular}{c|c|c} \hline
        	 & IoU of & Average \\
             & `person' (\%) & Detection Rate (\%) \\ \hline \hline
            Single-task \fukudaa{(VGG16-based PSPNet)} & \textbf{76.68} & 84.90 \\
            \fukudaa{Single-task cross-connected} & 76.24 & 85.03 \\ \hline
            Share1 & 75.23 & 88.00 \\
            Share2 & 75.31 & \textbf{88.10} \\
            Share3 & 75.47 & 87.56 \\
            Share4 & 75.39 & 87.76 \\ 
            Cross-stitch & 75.39 & 87.91 \\ \hline
            \fukudaa{Cross-connected (ours)} & 75.33 & 87.52 \\ \hline
        \end{tabular}
        }
        \label{tab:cityscapes_results}
    \end{center}
\end{table}
\begin{figure}[t]
	\vspace{-4mm}
    \begin{center}
		\includegraphics[width=\hsize]{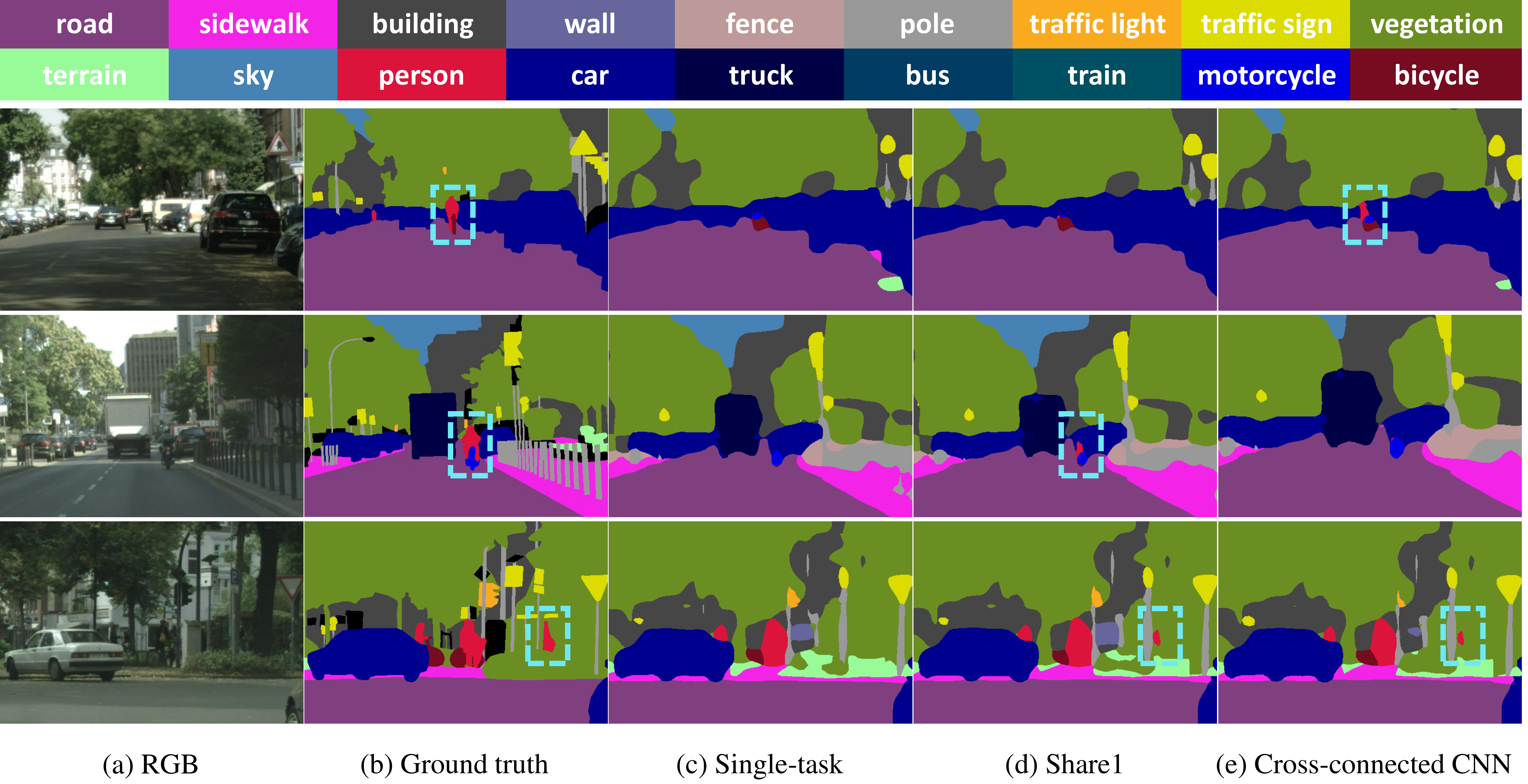}
	\end{center}
    \vspace{-5mm}
    \caption{Example segmentations on Cityscapes by baselines and cross-connected CNN. (d) Share1 and (e) Cross-connected CNN can detect more `person' regions. Detected `person' regions are highlighted by cyan dotted rectangles.}
	\label{fig:city_examples}
    \vspace{-7mm}
\end{figure}
\fukuda{Table \ref{tab:cityscapes_results} shows \allen{the} IoU and Detection Rate of `person' evaluated on Cityscapes. 
`Single-task' \allen{achieved} the highest IoU of 76.68\%\allen{,} and multi-task learning reduces IoU by about 1\%.}
\fukudaa{This is probably because \allen{the} bounding boxes have poor information on region shapes, \yoshia{and thus their contribution
to accurate segmentation, especially near region boundaries, is limited. 
In addition, the bounding boxes are provided only for  persons and not for backgrounds, which may decrease
semantic effects to dense pixel-wise labels.}
}

On the other hand, all multi-task CNNs \allen{achieved an} average \yoshia{Detection Rate} \allen{on} 2.62--3.20\% better than `Single-task' and 2.49--3.07\% than \allen{that of} `\fukudaa{Single-task cross-connected}'.
This means that multi-task learning with object detection enables \allen{the recognition of} more pixels within `person' regions.
\yoshia{Since there is only a difference \allen{of} less than 1\% between \allen{the} cross-connected CNN and multi-task baselines\yoshia{,
they} have \allen{a} \allen{near-equivalent} ability to leverage knowledge of pedestrian detection.}
\yoshia{This may be because a large portion of missed persons is occupied by unclear or low-resolution samples that are no longer detectable with visual information only.}
Fig. \ref{fig:city_examples} shows examples of semantic segmentation on the validation set.
The regions of `person' having low visibility are \fukudaa{detected} by multi-task CNNs (`Share1' and cross-connected CNN) but not by the single-task CNN.
Some instances can be \fukudaa{detected} only by multi-task baselines, others can be \fukudaa{detected} only by \fukudaa{the} cross-connected CNN. 
\yoshia{We include more result examples in the supplementary material. }

Comprehensively, \allen{the} cross-connected CNN has \allen{a} superior performance of pedestrian recognition, 
\allen{on the basis of } better performance in object detection and nearly equivalent performance in semantic segmentation.
\fukudaa{It is notable that cross-connected CNN benefits from multi-task learning of detection and segmentation even when they are trained on different datasets, although baselines deteriorate their detection performances.}

\subsection{Wild Bird Detection and Segmentation} 
Next, we evaluate the CNNs on the wild bird detection and segmentation, with which we aim to detect birds in landscape 
images around wind turbines to understand whether our method works for a different type of real scenes. 
In this experiment, we focus on \allen{the} `bird' class \yoshia{as a} 
target.
The goal is to improve the \yoshia{detection} 
performance of birds by \allen{the} multi-task learning of bird detection and segmentation of landscape images including birds.
Unlike the previous experiments, we train CNNs on the same dataset between the two tasks, so that
we can verify the generalization performance of bird detection with another dataset.\\

\noindent{\bf Datasets \hspace{2mm}}
We use two datasets constructed for wide-area surveillance of wild birds \cite{yoshihashi2015construction,takeki2016combining,trinh2016bird}: one for training and testing of object detection \cite{yoshihashi2015construction} and semantic segmentation \cite{takeki2016combining}, \allen{and} one only for testing of object detection \cite{trinh2016bird}.
We hereinafter refer to the first one as Dataset A and the second one as Dataset B.
Dataset A consists of 32,445 landscape images with ${5,616\times3,744}$ pixels taken under fine weather.
We use only the right half of them (${2,808\times3,744}$ pixels), which shows the surroundings of the wind turbine \yoshia{as in \cite{takeki2016combining}}.
In the experiments, we \allen{select} 138 images that have ground truths both for detection and segmentation.
46 and 113 birds taller than 15 pixels are used for training and testing in the experiments, respectively.
For segmentation, each pixel is annotated into 4 classes: bird, forest, sky, and wind turbine.
We use 40 images for training and 98 images for testing.
Due to the sparsity of their distribution, we set \allen{the} NMS threshold to 0.1.
Log-average Miss Rate is calculated on FPPI in ${[10^{-2}, 10^2]}$.

Dataset B is a set of images taken in a different power plant \allen{to those} from Dataset A.
Dataset B consists of 2,222 images with ${3,840\times2,160}$ pixels capturing landscapes under bad weather. 
Due to their complex backgrounds, it is \allen{a} more challenging dataset than Dataset A.
We \allen{select} 980 images where relatively more birds appear and use all of them for evaluating the generalization performance of bird detection.
615 birds taller than 15 pixels are evaluated in the test.
Evaluation conditions are the same as those of Dataset A.\\

\noindent{\bf Training Details \hspace{2mm}}
We use Dataset A for \allen{single-task} and multi-task learning of bird detection and segmentation.
Considering the low resolution of \allen{the} \yoshia{targets}, we remove 3 convolutional layers (conv4\_1-conv4\_3) and the 4th pooling layer from RPN.
RPN generates bounding boxes with a stride of 8 pixels in an input image.
\fukudaa{As in \ref{subsec:pedestrian_recognition},} we input images cropped with a size of ${512\times384}$ pixels.
We crop patches more around the wind turbine to avoid too many sky labeled patches in the training. 
Since Dataset A has ground truths for both tasks, we do not need to switch datasets during multi-task learning.
For one input patch, outputs of both tasks are simultaneously evaluated and all the network parameters are updated.\\

\noindent{\bf Results (Dataset A) \hspace{2mm}}
\begin{figure}[t]
	\vspace{-4mm}
	\begin{center}
		\includegraphics[width=0.8\hsize]{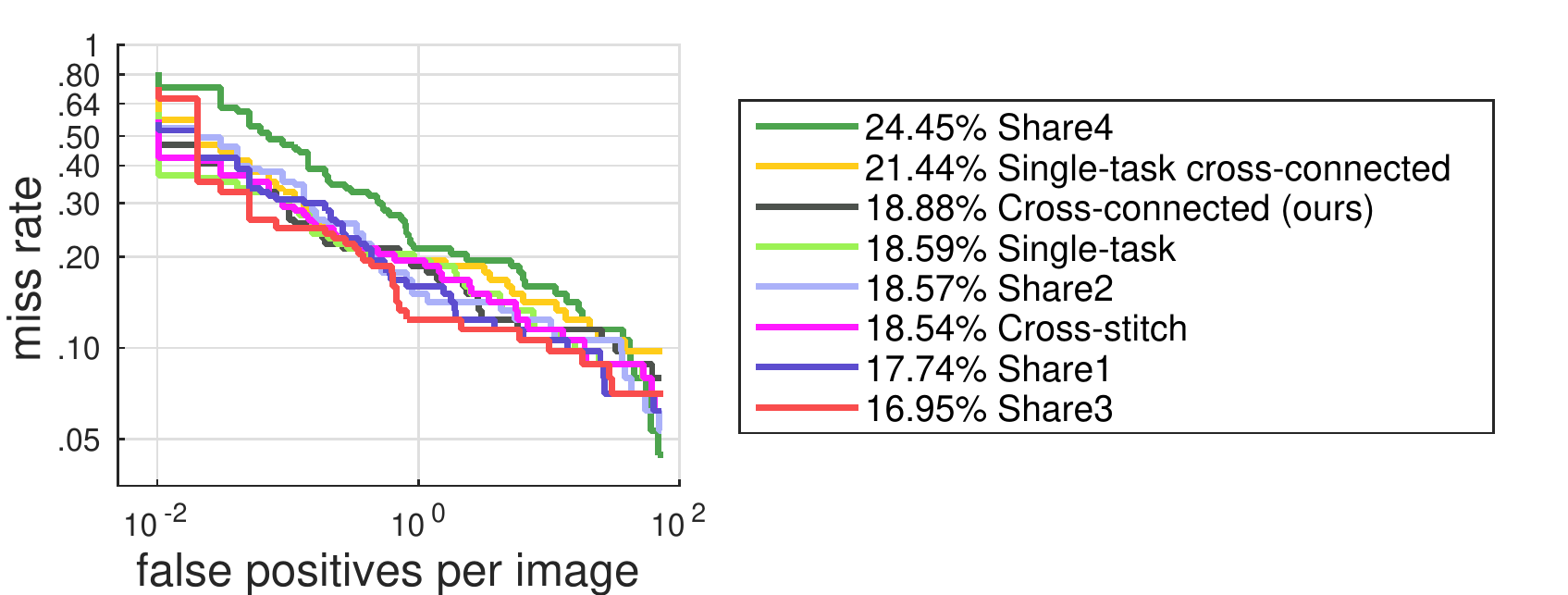}
	\end{center}
    \vspace{-5mm}
    \caption{Evaluation results of bird detection on Dataset A. The legend indicates log-average Miss Rate.}
	\label{fig:bird_det_a_results}
    \vspace{-4mm}
\end{figure}
\begin{table}[t]
	\begin{center}
    	\caption{\fukuda{Results of semantic segmentation on Dataset A}}
        {\tabcolsep = 4mm
    	\begin{tabular}{c|c|c} \hline
        	 & IoU of & Average \\
             & `bird' (\%) & Detection Rate (\%) \\ \hline \hline
            Single-task \fukudaa{(VGG16-based PSPNet)} & 33.89 & 59.39 \\
            \fukudaa{Single-task cross-connected} & 34.65 & 61.85 \\ \hline 
            Share1 & 34.07 & 61.16 \\
            Share2 & 34.39 & 61.65 \\
            Share3 & 34.78 & \textbf{62.83} \\
            Share4 & 31.74 & 54.67 \\ 
            Cross-stitch & 34.01 & 61.46 \\ \hline
            \fukudaa{Cross-connected (ours)} & \textbf{35.45} & 62.64 \\ \hline
        \end{tabular}
        }
        \label{tab:bird_seg_results}
        \vspace{-7mm}
    \end{center}
\end{table}
Fig. \ref{fig:bird_det_a_results} and Table \ref{tab:bird_seg_results} show \allen{the} evaluation results of bird detection and semantic segmentation on Dataset A\allen{,} respectively.
\fukudaa{In object detection, `Share3' achieves the best performance of 16.95\% among all methods.}
The cross-connected CNN has \allen{a} lower performance than all the baselines except `Share4' and `\fukudaa{Single-task cross-connected}', and shows almost the same result as `Single-task', `Share2'\fukuda{, and `Cross-stitch'}.
In the dataset, existing multi-task CNNs are enough to utilize knowledge of segmentation for detecting more birds.

In semantic segmentation, \allen{the} cross-connected CNN \allen{achieved} the highest IoU 35.45\%.
The Detection Rates of multi-task CNNs are better than the single-task CNN, which shows the benefit of multi-task learning.
\allen{The} cross-connected CNN \allen{improved} the Detection Rate by 3.25\% from \allen{that of} `Single-task' and \allen{outperforms} many baselines except `Share3'.
As the numerical values of IoU and Detection Rate in Table \ref{tab:bird_seg_results} are approximately correlated,
the improved IoU by multi-task learning should be because of more correctly detected instances rather than refinement of boundaries.
Regarding the Detection Rate, the cross-connected CNN has a similar performance to other multi-task baselines in leveraging knowledge from detection for segmentation.
More results are included in the supplementary material.

There are two possible reasons why the proposed CNN did not \allen{outperform} all baselines especially in detection.
First, it is relatively easy to \fukudaa{detect} the targets in Dataset A.
The background of images in Dataset A is usually occupied by a uniform blue sky; thus,
the performance may be saturated among existing models.
Second, we use the same dataset between the tasks during multi-task learning.
Since the layers in both tasks learn very similar representations, it could be easier to share knowledge than when using different datasets.
In such a case, \fukudaa{hard} parameter sharing could be sufficient for multi-task learning.\\

\noindent{\bf Results (Dataset B) \hspace{2mm}}
\begin{figure}[t]
	\vspace{-4mm}
	\begin{center}
		\includegraphics[width=0.8\hsize]{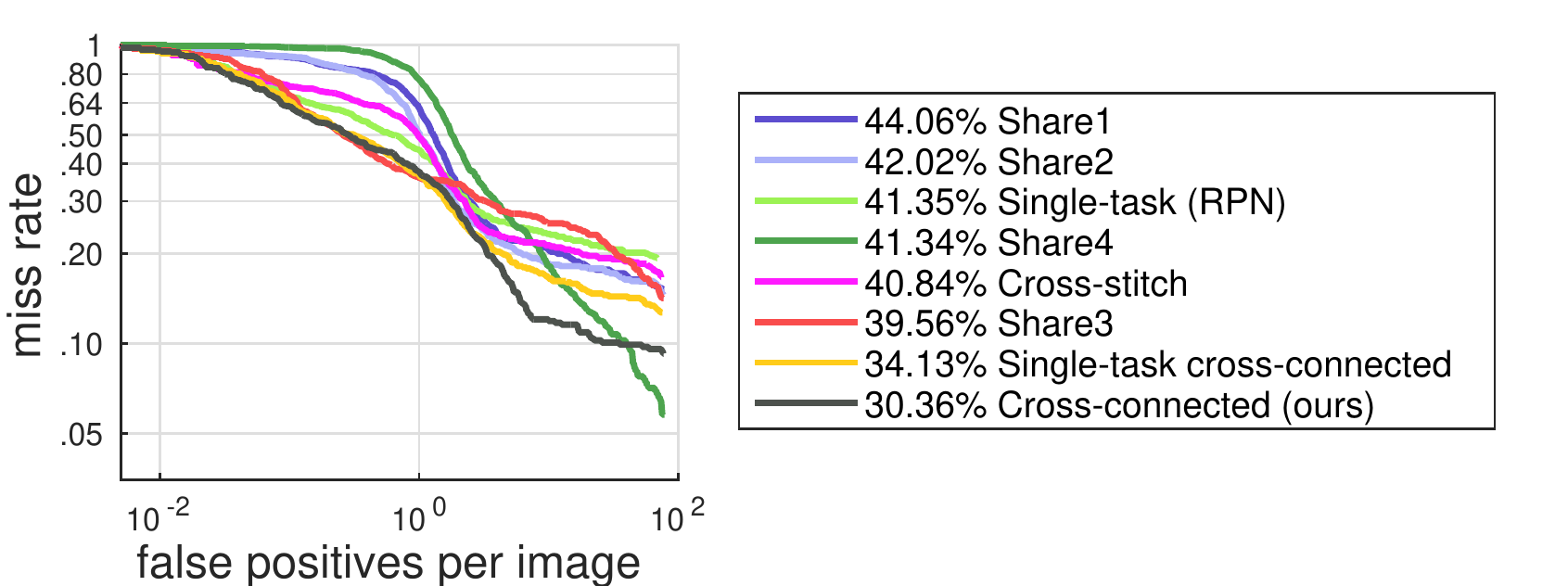}
	\end{center}
    \vspace{-5mm}
    \caption{Evaluation results of bird detection on Dataset B. The legend indicates log-average Miss Rate.}
	\label{fig:bird_det_b_results}
    \begin{center}
		\includegraphics[width=\hsize]{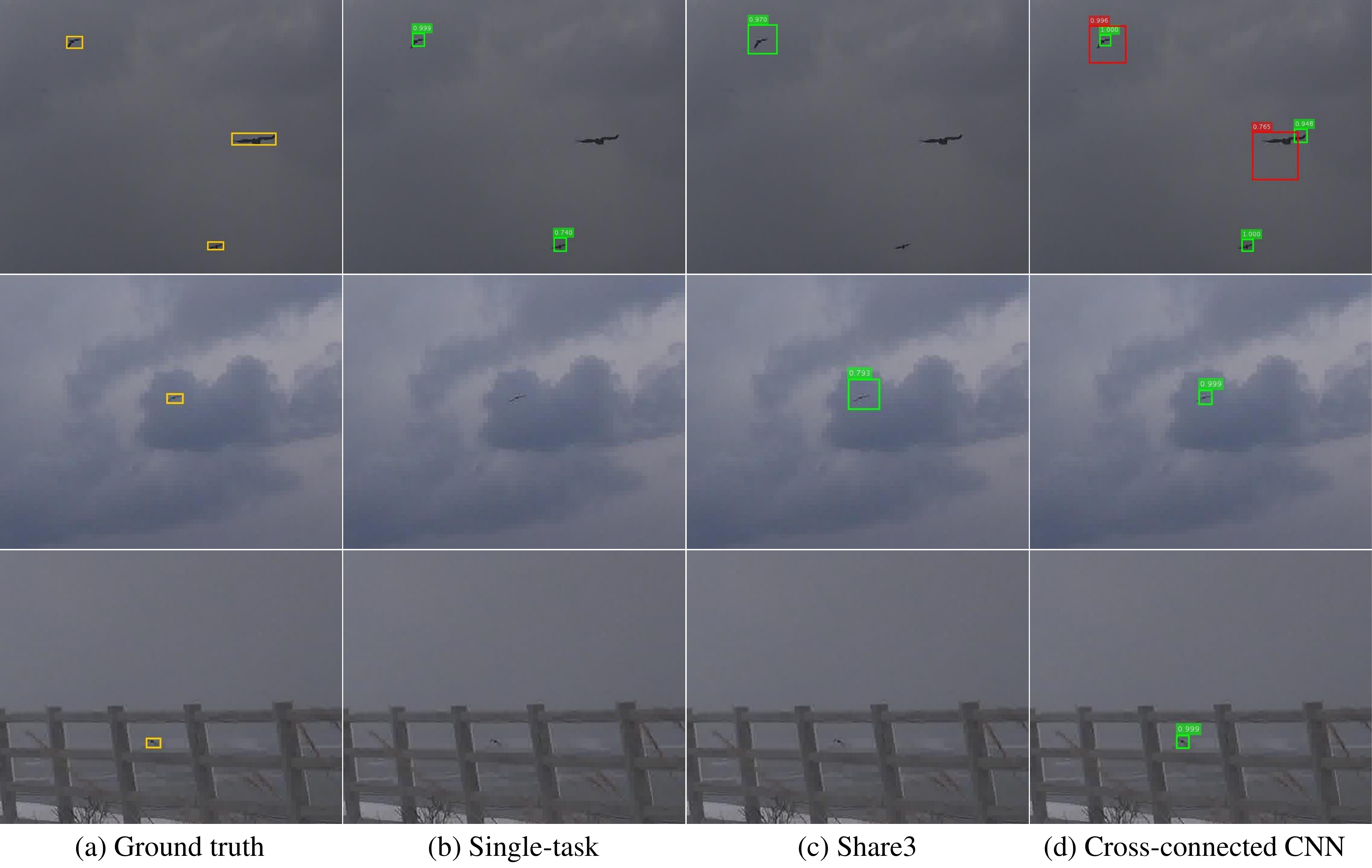}
	\end{center}
    \vspace{-5mm}
    \caption{Example detections on Dataset B \allen{with} baselines and cross-connected CNN. The ground truth, true positives, and false positives are displayed in yellow, green, and red, respectively.}
	\label{fig:bird_det_b_examples}
    \vspace{-5mm}
\end{figure}
Fig. \ref{fig:bird_det_b_results} shows the evaluation result of bird detection on Dataset B.
\allen{The} cross-connected CNN \allen{achieved a} 10.99\% better performance than \allen{that of} `Single-task' and \allen{a} 9.2\% better performance than \allen{that of} `Share3', the best of \allen{the} multi-task baselines.
This results suggests that \allen{the} cross-connected CNN obtains \allen{a} higher generalization performance than parameter sharing and cross-stitching.
The `single-task cross-connected' achieves 34.13\%, which is also better than all other baselines, probably due to more parameters than those of other baselines.
However, for the greatest improvement of the generalization performance, \fukudaa{multi-task learning with segmentation} is indispensable.
Fig. \ref{fig:bird_det_b_examples} shows example detections on Dataset B.
\allen{The} cross-connected CNN can detect more birds flying in the cloudy sky and in a small gap of the fence.

Together with the results on Dataset A, the baselines learn representations specialized only for Dataset A, but \allen{the} cross-connected CNN avoids \allen{overfitting to Dataset A} and learns more general representations.
This means that it takes less effort to prepare new training data when transferring a certain pre-trained model to \fukudaa{detection} in another \yoshia{location}.
This is a beneficial property, since it is laborious to manually annotate high-resolution landscape images.

\section{Conclusion}
\fukudaa{We have proposed a cross-connected CNN, a new multi-task CNN consisting of two inter-connected single-task CNNs.}
\fukudaa{In our architecture, two single-task streams pass their activation maps to each other via cross-connecting convolutional layers.}
\fukudaa{These layers enable activation maps to interact across their channels and learn how to utilize the knowledge obtained by task-wise pre-training.}
We \allen{evaluated} \fukudaa{our cross-connected CNN} \allen{using} a combination of object detection and semantic segmentation, \allen{and compared it} with \fukudaa{existing multi-task models}. 
\fukudaa{We conducted experiments on two datasets each of which targets pedestrians and wild birds.}
In pedestrian \fukudaa{detection and segmentation}, \fukudaa{we are the first to tackle} multi-task learning with different training datasets between \fukudaa{two tasks}.
We \allen{demonstrated} that our CNN outperforms baselines in detection performance, and leverages knowledge of segmentation.
In wild bird \fukudaa{detection and segmentation}, we \allen{demonstrated} that our CNN \fukudaa{acquires more general knowledge applicable to another dataset from limited training datasets.}
Future work will \allen{focus on \fukudaa{encouraging} a} more flexible \fukudaa{feature re-usage} by dense cross connections,
and \allen{the} application \allen{of it} to other combinations of tasks.

\bibliographystyle{splncs}
\bibliography{thebib}
\end{document}